\documentclass{article}

\usepackage{microtype}
\usepackage{graphicx}
\usepackage{subcaption}
\usepackage{booktabs}

\usepackage{hyperref}
\usepackage{amsmath}
\usepackage{amssymb}

\usepackage[accepted]{icml2022}

\icmltitlerunning{Approximate Bayesian Computation with Domain Expert in the Loop}

\newcommand{\yobs}{\mathbf y_{\mathrm{obs}}}
\newcommand{\sobs}{\mathbf s_{\mathrm{obs}}}
\newcommand{\ABC}{\text{ABC}}
\newcommand{\y}{\mathbf{y}}
\newcommand{\s}{\mathbf{s}}
\newcommand{\f}{\mathcal{F}}
\newcommand{\ggamma}{\boldsymbol{\gamma}}
\newcommand{\Nsim}{n_{\mathrm{sim}}}
\newcommand{\nobs}{n_{\mathrm{obs}}}

\DeclareMathOperator*{\argmax}{arg\,max}

\begin{document}

\twocolumn[
\icmltitle{Approximate Bayesian Computation with Domain Expert in the Loop}

\begin{icmlauthorlist}
\icmlauthor{Ayush Bharti}{aalto}
\icmlauthor{Louis Filstroff}{aalto}
\icmlauthor{Samuel Kaski}{aalto,manchester}
\end{icmlauthorlist}

\icmlaffiliation{aalto}{Department of Computer Science, Aalto University, Espoo, Finland}
\icmlaffiliation{manchester}{Department of Computer Science, University of Manchester, Manchester, United Kingdom}

\icmlcorrespondingauthor{Ayush Bharti}{ayush.bharti@aalto.fi}

\icmlkeywords{Machine Learning, ICML, Approximate Bayesian Computation, Model Misspecification, Human-in-the-loop}

\vskip 0.3in
]

\printAffiliationsAndNotice{} 

\begin{abstract}

Approximate Bayesian computation (ABC) is a popular likelihood-free inference method for models with intractable likelihood functions. As ABC methods usually rely on comparing summary statistics of observed and simulated data, the choice of the statistics is crucial. This choice involves a trade-off between loss of information and dimensionality reduction, and is often determined based on domain knowledge. However, handcrafting and selecting suitable statistics is a laborious task involving multiple trial-and-error steps. In this work, we introduce an active learning method for ABC statistics selection which reduces the domain expert's work considerably. By involving the experts, we are able to handle misspecified models, unlike the existing dimension reduction methods. Moreover, empirical results show better posterior estimates than with existing methods, when the simulation budget is limited.

\end{abstract}

\section{Introduction}

Likelihood-free inference has considerably extended the applicability domain of probabilistic inference, to the set of problems where a simulator is available even though the likelihood function is not known or feasibly computable. Approximate Bayesian computation (ABC) \cite{Marin2011, Lintusaari2016, Sisson2018, Beaumont2019} has emerged as a popular method for likelihood-free inference in a number of fields such as population genetics \cite{Pritchard1999,Beaumont2010}, cosmology \cite{Akeret_2015}, and radio propagation \cite{Bharti2021}, among others. ABC permits sampling from an approximate posterior distribution of a generative model by comparing summary statistics of simulated and observed (high-dimensional) data. However, the success of ABC may hide from view the fact that major problems are still only partially solved. In this paper, we discuss the problem of choosing summary statistics, a key part of ABC which is often missed in clean theoretical works and only treated case-specifically in concrete inference studies.

Recent works have proposed to circumvent choosing statistics by learning a suitable representation from data with neural networks \citep{Papamakarios2016, Lueckmann2016, Lueckmann2019, Izbicki2019}. However, training neural networks requires a large amount of data, which means extensive simulator runs in the ABC context. Thus, in a \emph{low-simulation regime}, where the number of available simulations is limited, choosing the summary statistics is unavoidable -- and useful in any case. This choice involves navigating a difficult trade-off between: 1) information loss due to data summarization and hence lower-quality posterior approximations, and 2) curse of dimensionality, requiring exponentially increasing numbers of simulator runs. A low-dimensional set of statistics which is highly informative about the model parameters would be ideal, but obtaining such a set is non-trivial. The only way out of this conundrum is to bring in additional domain knowledge, and hence practitioners end up spending a large proportion of the time of their likelihood-free inference projects in choosing suitable statistics.

Several methods have been proposed to automatically reduce the dimension of a given set (or pool) of available summary statistics to use in an ABC method, see \citet{Blum2013, prangle2015summary} for exhaustive surveys. These include methods based on subset selection \citep{Joyce2008, Nunes2010,Blum2010, Barnes2012, Blum2013}, projection \citep{ Wegmann2009, Fearnhead2012, Aeschbacher2012, Jiang2017, chen2021neural}, and regression adjustment \citep{Beaumont2002, Blum2010nonlinear, Bi2021}. However, none of these methods are able to handle low-simulation regimes and model misspecification, which occurs when there is a mismatch between the simulator and the true data-generating mechanism. Under model misspecification, dimension-reducing ABC methods may produce summary statistics which will never replicate the observed value irrespective of the parameter setting, which in turn causes problems in the ABC \cite{Frazier2020}. We call those statistics misspecified. In low-simulation regimes, on the other hand, these methods would become susceptible to fitting to noisy, uninformative statistics \cite{Blum2013}. Therefore, existing methods cannot alone offer a sufficient solution to the statistics selection problem.

In practice, domain knowledge is brought in likelihood-free inference by experts handcrafting and selecting the statistics manually. This is necessary, as the choice depends on the model, data and application at hand, albeit laborious, as it involves multiple trial-and-error steps. In this paper, we propose a human-in-the-loop ABC statistics selection method which considerably eases the work of domain experts, extending the statistics selection method of \citet{Barnes2012}. Taking the regression-based ABC methods as a case study, we show that by including the experts in the inference loop, we achieve better posterior characterization when the model is misspecified or when model evaluation is costly. We assume that expert knowledge is tacit, that is, the expert cannot easily produce an optimal set of informative statistics, but can recognize a good statistic when presented with it. Additionally, the expert can recognize potentially misspecified statistics and exclude them. We adopt a sequential Bayesian experimental design (BED) \citep{chaloner1995bayesian,ryan2016review} approach to sequentially select the most informative statistics to present to the expert using a forward-stepwise selection method \cite{Hastie2009}. To the best of our knowledge, domain experts have not been formally involved in ABC methods so far. We show clearly better empirical performance than with existing methods on two models with intractable likelihoods: a quantile distribution and a radio propagation model.

\section{Basics \& Motivation}\label{sec:background}
We introduce some basics on ABC methods in Section~\ref{sec:abc} and demonstrate their potential pitfalls in Section~\ref{sec:motivation}.

\subsection{Approximate Bayesian computation}\label{sec:abc}

Let $\mathcal{Y}$ be the data space and $\mathcal{M}_\Theta = \{\mathbb{P}_\theta: \theta \in \Theta \subset \mathbb{R}^q\}$ a parametric model family of distributions $\{\mathbb{P}_\theta\}$ on $\mathcal{Y}$. We assume that $\mathbb{P}_\theta$ does not have a tractable likelihood function given observed data $\yobs$ (comprising $\nobs$ samples), but it is possible to simulate independent and identically distributed (i.i.d.) samples from $\mathcal{M}_\Theta$ given some $\theta$. For such models, ABC methods can be used to approximate their posterior distribution $p(\theta| \yobs) \propto p(\yobs | \theta ) p(\theta)$ in a non-parametric manner, where $p(\theta)$ denotes the prior beliefs and $p(\yobs|\theta)$ is the joint likelihood function. We now briefly describe some of the basic ABC methods.

\paragraph{Rejection-ABC} Consider a deterministic function $\eta$ mapping the data to a lower-dimensional space of summary statistics such that $\sobs = \eta(\yobs)$ is the vector of summary statistics of $\yobs$. The basic rejection-ABC algorithm \cite{Pritchard1999} proceeds in the following manner: (1) Sample $\theta^* \sim p(\theta)$; (2) Simulate $\y^* \sim \mathbb{P}_{\theta^*}$ and compute $\s^* = \eta(\y^*)$; (3) If $\varrho(\s^*, \sobs) < \epsilon$, accept $\theta^*$. Here $\varrho(\cdot, \cdot)$ is a distance function and $\epsilon$ is a tolerance threshold. Repeating the algorithm yields a set $\{\theta_i\}_{i=1}^n$ of accepted parameter values which are i.i.d. samples from the approximate posterior $p(\theta | \varrho(\s, \sobs) < \epsilon) \approx p(\theta | \yobs)$. For a fixed simulation budget of $n_{\mathrm{sim}}$ samples, a practical solution is to specify the tolerance as the ratio $\epsilon=n_\epsilon/ \Nsim$, where $n_\epsilon$ is the number of accepted samples out of $\Nsim$.

\paragraph{Regression-ABC} Regression adjustment approaches to ABC \cite{Blum2017} aim to account for the difference between the simulated and observed statistic by adjusting the parameter values. Given samples $(\theta_i, \s_i)_{i=1}^{n_\epsilon}$ obtained with rejection-ABC, a homoscedastic regression model,
\begin{equation}
    \theta_i = \varphi(\s_i) + \varepsilon_i, \quad i=1,\dots,n_\epsilon,
\end{equation}
is fitted in the vicinity of $\sobs$, where $\varphi(\cdot)$ is the conditional expectation $\mathbb{E}[\theta|\s]$, and $\varepsilon_i$ are the residuals. The parameter samples are then adjusted as $\tilde \theta_i = \hat \varphi(\sobs) + \hat \varepsilon_i$, with $\hat \varphi$ being the estimate of $\mathbb{E}[\theta|\s]$ and $\hat \varepsilon_i$ being the $i^\textup{th}$ empirical residual. \citet{Beaumont2002} assumed $\varphi$ to be linear, while it was later extended to heteroscedastic non-linear adjustment in \citet{Blum2010nonlinear}. \citet{Blum2013} proposed a regularized version of the linear-ABC method via ridge regression. We refer to these methods as linear-ABC, neural-ABC, and ridge-ABC, respectively.

\begin{figure}[t]
\includegraphics[width=\columnwidth]{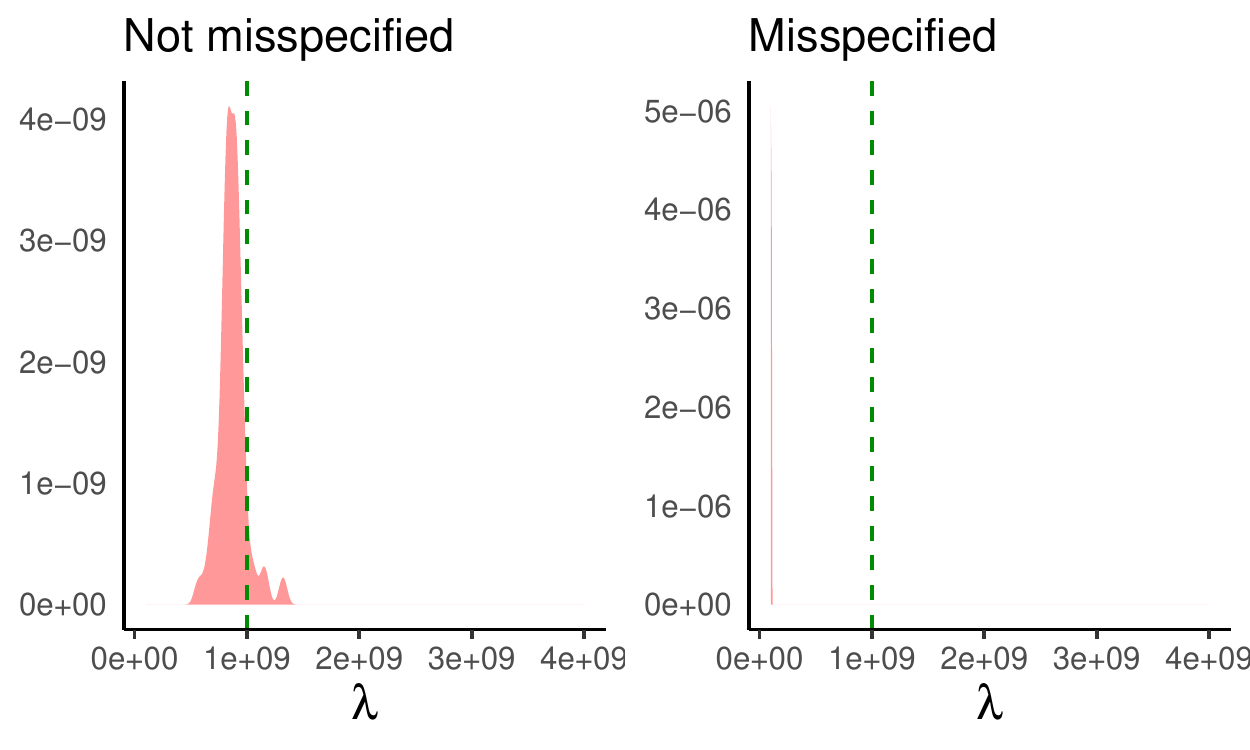}
\caption{Catastrophic performance of linear-ABC under model misspecification for the $\lambda$ parameter of the radio propagation model, see Section~\ref{sec:shotNoise} for experiment details. The dashed green line denotes the true parameter value. We observe that the ABC samples concentrate far away from the true value, on the left prior boundary, when the model is misspecified.}
\label{fig:motivation_misspec}
\vskip -0.2in
\end{figure}
\begin{figure}[t]
\begin{center}
\centerline{\includegraphics[width=\columnwidth]{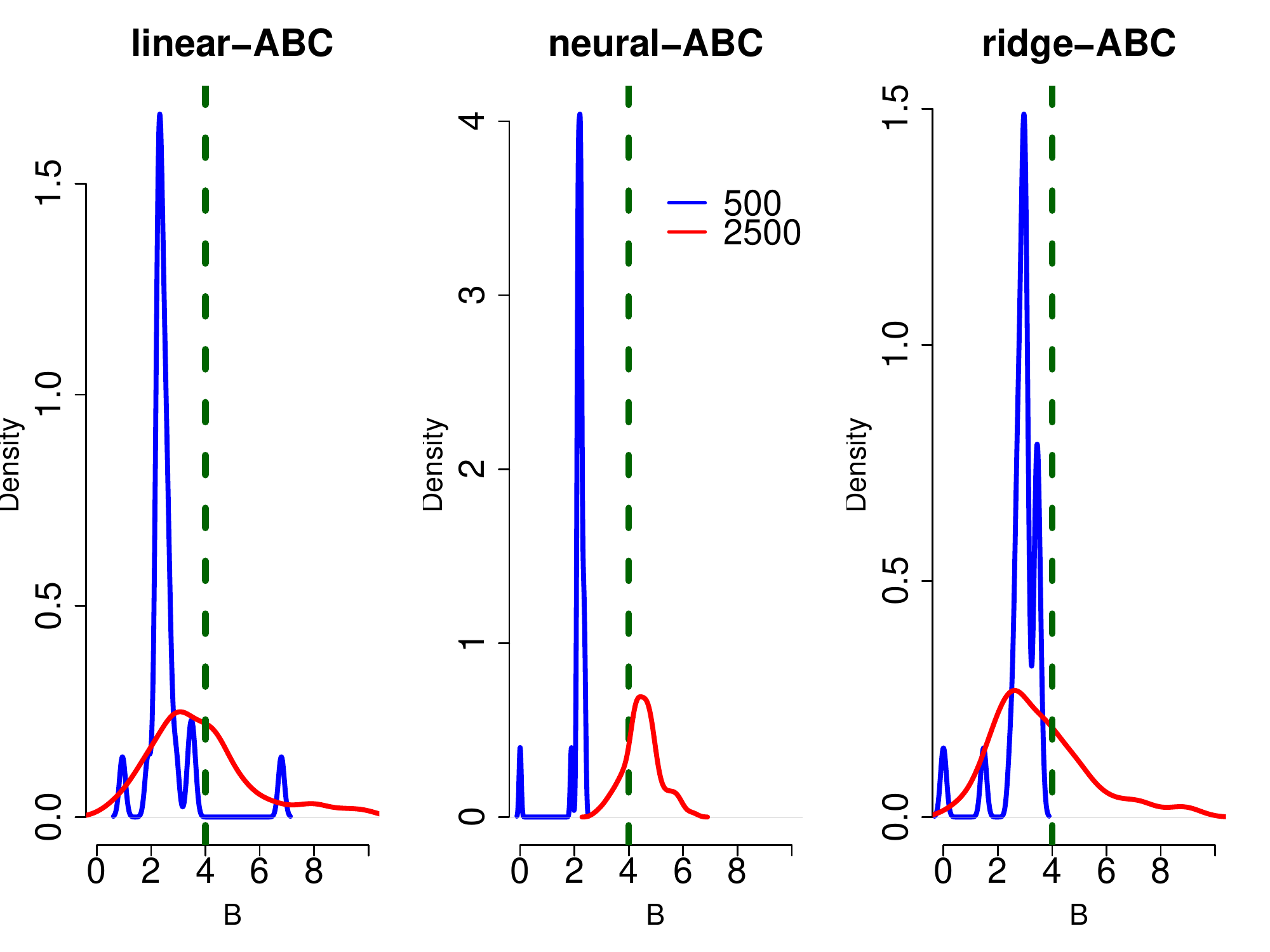}}
\caption{Inability of regression-ABC methods to handle low-simulation regime. The ABC posteriors obtained from linear-ABC, neural-ABC, and ridge-ABC for the parameter $B$ of the g-and-k distribution (see Section~\ref{sec:gk} for details) using $\Nsim = 500$ (blue) and $\Nsim = 2500$ (red) simulated samples with $\epsilon=5\%$. The dashed green line denotes the true parameter value. The regression layer in these methods fits to noisy, uninformative statistics, thereby leading to ABC posteriors (blue curves) concentrated away from the true value.}
\label{fig:motivation_lowsim}
\end{center}
\end{figure}

\subsection{Pitfalls of regression-ABC methods} \label{sec:motivation}

\paragraph{Model misspecification} The regression-based methods have been shown to concentrate the ABC posterior around the true value when the model is correctly specified \cite{Li2018}. However, under model misspecification, i.e., when the true data-generating mechanism does not belong to $\mathcal{M}_\Theta$, they can concentrate the ABC posterior on a completely different region of the parameter space than rejection-ABC, which can be outside the prior range for bounded parameters \cite{Frazier2020}. This phenomenon is demonstrated in Fig.~\ref{fig:motivation_misspec} for the radio propagation model (see Section~\ref{sec:shotNoise} for details) using linear-ABC. When the model is misspecified, we see the ABC posterior concentrates on the prior boundary, far away from the true parameter value. This occurs when just one of the statistics is misspecified, as is the case in Fig.~\ref{fig:motivation_misspec}. In this paper, we utilize the fact that the expert will be able to detect this behaviour and exclude potentially misspecified or out-of-distribution statistics from being included in the regression-ABC methods.

\paragraph{Low-simulation regime} Inability of the regression-based methods to handle the low-simulation regime is exemplified by the g-and-k distribution (see Section~\ref{sec:gk}) in Fig.~\ref{fig:motivation_lowsim}. We observe that for small $\Nsim$, the resulting ABC posteriors can get concentrated away from the true parameter value. As there are few samples to perform the least-squares fit in regression-ABC methods, they may overfit to uninformative statistics. In such cases, these methods may over-adjust the parameter values in the direction of such noisy statistics \cite{Blum2013}.

\section{The Method}\label{sec:methodology}

We propose including in the statistics selection loop a domain expert, who will be able to assess which statistics would be useful or misspecified, and who currently needs to do that choice completely manually. The expert may evaluate the usefulness of a given statistic by, e.g., checking relevant literature. As this is laborious, we would not want to repeat it for all possible candidates. In this section, we introduce an experimental design approach which helps reduce the expert's effort. We formulate their feedback as a probabilistic modelling problem as well, with the knowledge of the expert as a latent variable, as described in Section~\ref{sec:feedbackModel}. This turns querying the expert into an automatic experimental design problem, which is presented in Section~\ref{sec:utility}. Section~\ref{sec:setting} describes the problem set-up, and Alg.~\ref{alg:final} outlines the proposed human-in-the-loop (HITL) ABC algorithm. 

\subsection{Setting}\label{sec:setting}
Consider a finite pool of candidate summary statistics $\mathcal{S} = \{s_1, s_2, \dots, s_w\}$ available for ABC. We introduce a binary variable $\gamma_j \in \{0,1\}$ to indicate the inclusion or exclusion of the statistic $s_j \in \mathcal{S}$, $j=1,\dots,w$ to the summarizing function $\eta(\cdot)$. Denote by $\boldsymbol{\gamma} = [\gamma_1, \dots, \gamma_w]^\top$ the binary vector corresponding to a vector of statistics $\s = \eta(\mathbf y)$, such that $\gamma_j = 1$ implies $s_j$ is an element of $\s$. We denote the approximate posterior obtained by applying an ABC method with tolerance $\epsilon$ by $p^\epsilon_\ABC(\theta | \yobs, \ggamma)$. For $\ggamma = \mathbf{0}$, we set $p^\epsilon_\ABC(\theta | \yobs, \ggamma) = p(\theta)$. Let $\ggamma^*$ represent the desired subset\footnote{Here, the desired subset of statistics is understood as the subset that achieves the optimal trade-off between minimum dimensionality and information regarding the parameters.} of statistics $\s^* \subset \mathcal{S}$ to be used in the ABC method such that $p^\epsilon_\ABC(\theta | \yobs, \ggamma^*)$ is the target ABC posterior. We query the expert $\mathcal{E}$ regarding the elements of $\mathcal{S}$ with the goal to converge towards $\ggamma^*$ as quickly as possible. We assume that the expert is queried only once about a given statistic, and that querying the expert is costly.

\subsection{Expert feedback model}\label{sec:feedbackModel}

We assume that the expert provides binary feedback $f_j \in \{0,1\}$ regarding the relevance of the $j^\textup{th}$ statistic $s_j$ and interpret the answer as feedback about $\gamma_j$. More precisely, we model $f_j$ as a noisy version of $\gamma_j$, such that
\begin{align}
    \gamma_j & \sim \text{Bernoulli}(\rho_j), \\
    f_j | \gamma_j & \sim \gamma_j \text{Bernoulli}(\pi) + (1-\gamma_j)\text{Bernoulli}(1-\pi). \label{eq:feedbackModel}
\end{align}
The hyperparameter $\pi \in [0,1]$ quantifies the level of noise or uncertainty in the feedback, i.e., we have $f_j = \gamma_j$ with probability $\pi$. Of course, the method is intended to work with domain experts having prior knowledge about the statistics, for whom $\pi$ would be close to 1. The hyperparameter $\rho_j$ corresponds to the prior probability of the $j^\textup{th}$ statistic being included. This model was first proposed by \citet{Daee2017} to get feedback on the relevance of features to be used in regression, and later applied to precision medicine by \citet{Sundin2018}. Note that the feedback is independent of $\yobs$.

By marginalization, it is straightforward to show that the feedback $f_j$ is a Bernoulli random variable with probability of success $\omega_j = \pi \rho_j + (1-\pi)(1-\rho_j)$. We can further characterize the posterior probability of $\gamma_j$ given $f_j$ as a Bernoulli distribution of parameter $\nu_j$, where
\begin{equation}
    \nu_j = \frac{\pi^{f_j} (1-\pi)^{1-f_j}\rho_j}{\omega_j^{f_j}(1-\omega_j)^{1-f_j}}.
\end{equation}
For simplicity, we assume $\rho_j = \rho$ for all $j$ in the remainder of the paper.

Denote by $\mathcal{J} = \{j_1, j_2, \dots,j_m\}$ the indices of the $m\leq w$ summary statistics that have been queried from the expert. The corresponding feedback sequence is denoted as $\mathcal{F} = \{f_{j_1}, f_{j_2}, \dots, f_{j_m}\}$.  Then the posterior $p(\ggamma|\f)$ is 
\begin{equation}
    p(\ggamma|\f) = \prod_{j \in \mathcal{J}} p(\gamma_j|f_j) \prod_{j \notin {\mathcal J}} p(\gamma_j),
\end{equation}
where $j \notin \mathcal{J}_k$ denotes the indices of statistics for which feedback has not been queried yet.
\paragraph{ABC posterior based on feedback} Given that we observe expert feedback $\mathcal{F}$ and not $\ggamma$, we define the ABC posterior based on $\mathcal{F}$ as
\begin{equation}\label{eq:feedbackABC}
    p_\ABC^\epsilon(\theta| \yobs, \f) := \hspace{-1ex} \sum_{\ggamma \in \{0,1\}^w} p_\ABC^\epsilon(\theta|\yobs, \ggamma) p(\ggamma|\f),
\end{equation}
that is, we integrate out our current beliefs about $\ggamma$. Note that $p_\ABC^\epsilon(\theta|\yobs, \ggamma)$ does not have a closed-form expression. Nonetheless, it is possible to obtain i.i.d. samples $\theta^{(i)}$ from $p_\ABC^\epsilon(\theta| \yobs, \mathcal{F})$ in the following manner: 
\begin{enumerate}
    \item Sample $\ggamma^{(i)} \sim p(\ggamma|\f)$;
    \item Sample $\theta^{(i)} | \ggamma^{(i)} \sim p_\ABC^\epsilon(\theta|\yobs, \ggamma^{(i)})$. 
\end{enumerate}
As querying the expert is costly, we want $p_\ABC^\epsilon(\theta | \yobs, \f)$ to converge towards $p_\ABC^\epsilon(\theta | \yobs, \ggamma^*)$ with the least amount of feedback.

\subsection{Sequential experimental design}\label{sec:utility}

We design a sequential Bayesian experiment to select the next statistic to get feedback on from the expert. We refer the reader to Appendix~\ref{app:bed} for background on Bayesian experimental design. Our utility function is the expected KL divergence between the ABC posteriors (defined in Eq.~\eqref{eq:feedbackABC}) before and after receiving a new feedback. Denote by $\f_k$ the feedback collected after iteration $k$, and by $\mathcal{J}_k$ the indices of the queried statistics (in particular, $\f_0 = \emptyset$ and $\mathcal{J}_0 = \emptyset$). Thus, at iteration $k+1$, the utility maximizing statistic $s_{j^*}$ with index
\begin{equation}\label{eq:optProb}
    j^* = \underset{j \notin \mathcal{J}_k}{\argmax} ~ U_{k+1}(j), 
\end{equation}
is chosen. The utility function reads
\begin{equation}\label{eq:utility}
    U_{k+1}(j) = \mathbb{E}_{p(\tilde f_j| \f_k)} \left[ \mathfrak{D}_{k}^{\text{KL}}(\tilde f_j)\right], \quad \text{where}
\end{equation}
$$\mathfrak{D}_{k}^{\text{KL}}(\tilde f_j) = \text{KL} [ p_\ABC^\epsilon(\theta | \yobs, \f_k, \tilde f_j) \: || \; p_\ABC^\epsilon(\theta | \yobs, \f_k)].$$

The expectation in Eq.~\eqref{eq:utility} is taken w.r.t. the posterior predictive distribution $p(\tilde f_j|\f_k)$, as the feedback is only observed after actually querying the expert. Recall that in our setting, the expert can only be queried once about each statistic $s_j$, leading to feedback $f_j$. Also, $f_j$ is independent of $f_j'$ for $j \neq j'$. It follows that $p(f_j|\f_k) = p(f_j)$. Therefore, we can further write Eq.~\eqref{eq:utility} as the Bernoulli expectation
\begin{align}
    U_{k+1}(j) & = \Pr(\tilde f_j = 1) \mathfrak{D}_{k}^{\text{KL}}(\tilde f_j = 1) \notag \\
    & + \Pr(\tilde f_j = 0) \mathfrak{D}_{k}^{\text{KL}}(\tilde f_j = 0).
\end{align}
\begin{algorithm}[t]
  \caption{Human-in-the-loop (HITL) ABC}
  \label{alg:final}
\begin{algorithmic}
  \STATE {\bfseries Input:} data $\yobs$, model $\mathcal{M}_\Theta$, expert $\mathcal{E}$, prior $p(\theta)$, pool $\mathcal{S}$, tolerance $\epsilon$, stopping threshold $\delta$
  \REPEAT
    \STATE Sample $\{ \theta_{k}^{(i)} \}_{i=1}^n \sim p_{\ABC}(\theta | \yobs, \mathcal{F}_{k})$ (see Section~\ref{sec:feedbackModel})
  \FOR{$j \notin \mathcal{J}_k$}
  \STATE $\{ \theta_{k+1}^{(i)} \}_{i=1}^n \sim p_{\ABC}(\theta | \yobs, \mathcal{F}_{k}, \tilde f_j)$ for $\tilde f_j = \{0,1\}$
  \STATE Compute utility $U_{k+1}(j)$ from \eqref{eq:utility}
  \ENDFOR
  \STATE Find $j^*$ by solving \eqref{eq:optProb} 
  \STATE Query $s_{j^*}$ from the expert to get feedback $f_{j^*}$
  \STATE $\mathcal{F}_{k+1} = \mathcal{F}_{k} \cup f_{j^*} $ 
  \UNTIL{stopping criterion is met}
  \STATE \textbf{Output:} 
  ABC posterior $p_\ABC^\epsilon(\theta | \yobs, \hat \ggamma)$, where $\hat \ggamma$ is given by \eqref{eq:output}   
\end{algorithmic}
\end{algorithm}

Given i.i.d. samples $\{\theta^{(i)}_{k+1}\}_{i=1}^n \sim p_{\ABC}(\theta | \yobs, \f_k, \tilde f_j)$ and $\{\theta^{(i)}_{k}\}_{i=1}^n \sim p_{\ABC}(\theta | \yobs, \f_{k})$, we estimate the KL divergence using the 1-nearest neighbour density estimate \cite{Wang2006, Jiang2018}
\begin{equation}\label{eq:KL_estimator}
    \mathfrak{D}_{k}^{\text{KL}}(f_j) \approx \frac{q}{n} \sum_{i=1}^n \log \frac{\mathrm{min}_j \Vert \theta^{(i)}_k - \theta^{(j)}_{k-1} \Vert }{\mathrm{min}_{j \neq i}^n \Vert \theta^{(i)}_k - \theta^{(j)}_k \Vert} + \log \frac{n}{n-1},
\end{equation}

which guarantees almost-sure convergence to the true divergence \cite{Perez-Cruz2008}. Moreover, the estimator has a time complexity of $\mathcal{O}(n \log n)$ \cite{Jiang2018}, and thus scales well with the number of samples. We use k-d trees \cite{Bentley1975, Songrit2001} to implement it with $n = 4000$ samples for all the experiments in this paper.

\paragraph{Stopping criterion} We stop Alg.~\ref{alg:final} as soon as the utility of the remaining statistics falls below a pre-defined threshold $\delta$, i.e. $U_{k+1}(j) \leq \delta$ for $j \notin \mathcal{J}_k$. Addition of any new statistic from this stage onwards barely impacts the ABC posterior, indicating the absence of informative statistics in the remaining pool. We therefore assign $\gamma = 0$ to the statistics not queried before stopping the algorithm. To set the value of $\delta$, we follow the argument by \citet{Barnes2012} and pick $\delta$ which is larger than the estimated KL divergence between samples from the same distribution. Formally, let $\mathbf{X} = (X^{(1)}, \dots, X^{(n)})$ be a sample from a $q$-dimensional density $p_X$. We sample $\{\mathbf X_i\}_{i=1}^M \sim p_X$ and set $\delta = \max_{i,j}~ \text{KL}(\mathbf{X}_i, \mathbf{X}_j)$,  for  $i,j = 1, \dots, M$. 

\paragraph{Output of the algorithm} At the end of each iteration $k$, the current estimate of the statistics indicator vector is $\hat \ggamma_k = (\hat{\gamma}_{k,1}, \dots, \hat{\gamma}_{k,w})$, where  
\begin{equation}\label{eq:output}
    \hat{\gamma}_{k,j} = \begin{cases}
    \underset{\gamma_j \in \{ 0, 1 \} }{\argmax}~ p(\gamma_j|f_j), & \text{if}~ j \in \mathcal{J}_k\\
    0,              & \text{otherwise}.
\end{cases} 
\end{equation}
The final output of Alg.~\ref{alg:final} is the ABC posterior $p_\ABC^\epsilon(\theta | \yobs, \hat \ggamma)$ where $\hat \ggamma$ is obtained from Eq.~\eqref{eq:output} given all the collected feedback.

\section{Experiments} \label{sec:experiments}

In this section, we empirically assess the performance of the proposed HITL-ABC method against regression-ABC methods under model misspecification in Section~\ref{sec:shotNoise}, and in low-simulation regimes in Section~\ref{sec:gk}. Lastly, the sensitivity to hyperparameters is analyzed in Section~\ref{sec:hyperparameter}. The source code is available at \url{https://github.com/lfilstro/HITL-ABC}.

\paragraph{Implementation details} Our algorithm can be implemented on top of any ABC method. To identify the effect of the novel contribution, we choose the same method used as a baseline method in comparisons, namely the regression adjustment approach of \citet{Beaumont2002} (linear-ABC). The regression-ABC methods are implemented using the \texttt{abc} R package \cite{Csillery2012}. For all the experiments, we assume bounded uniform priors on the parameters and use a logit transform \cite{Blum2010nonlinear} before adjusting them to ensure adjusted parameters do not fall outside the prior range. The statistics are normalized by an estimate of their mean absolute deviation before computing the distance to account for the difference in magnitudes. The confidence in the feedback is set to $\pi=0.95$, and the stopping criterion is $\delta = 0.06$. Assuming each statistic is equally likely to be included or excluded \textit{a priori}, we set $\rho = 0.5$. We assume $\varrho(\cdot, \cdot)$ to be the Euclidean norm $\Vert \cdot \Vert$, as is a typical choice in ABC. Lastly, a run of the algorithm uses the same simulated data at each iteration for computational ease.

\begin{figure*}[t]
    \begin{center}
        \begin{tabular}{ccc}
            \includegraphics[width=0.31\linewidth]{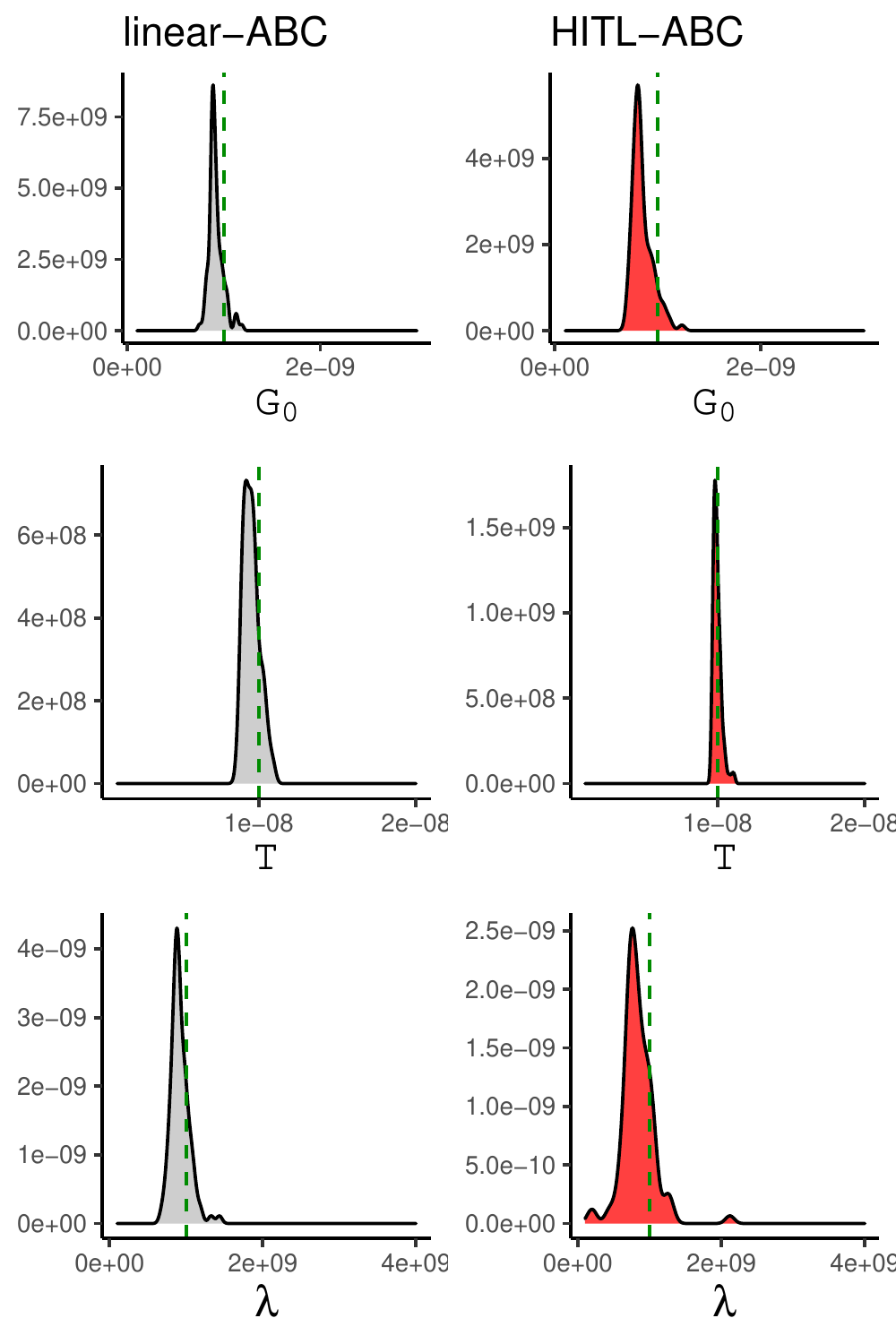} & 
            \includegraphics[width=0.31\linewidth]{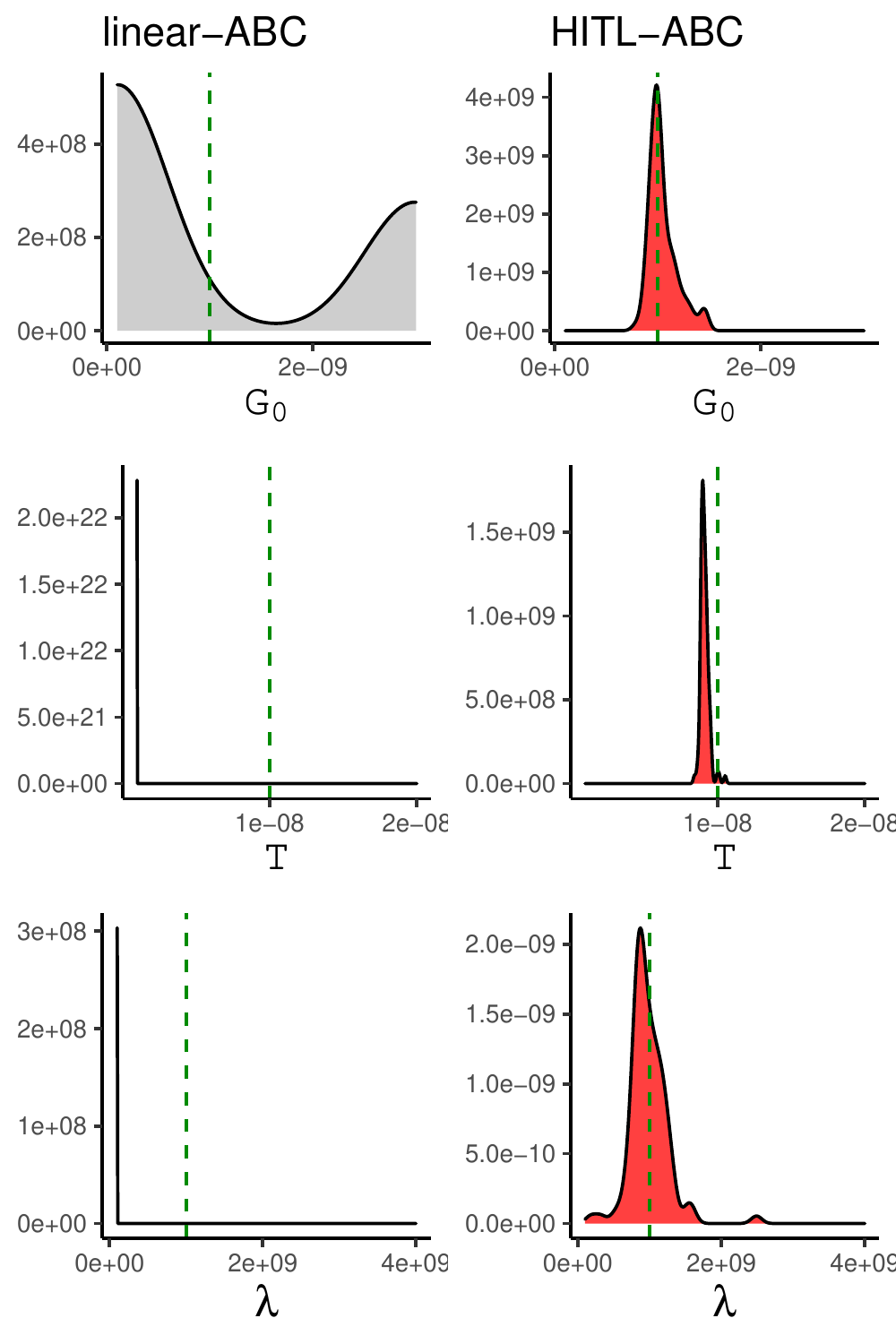} & 
            \includegraphics[width=0.31\linewidth]{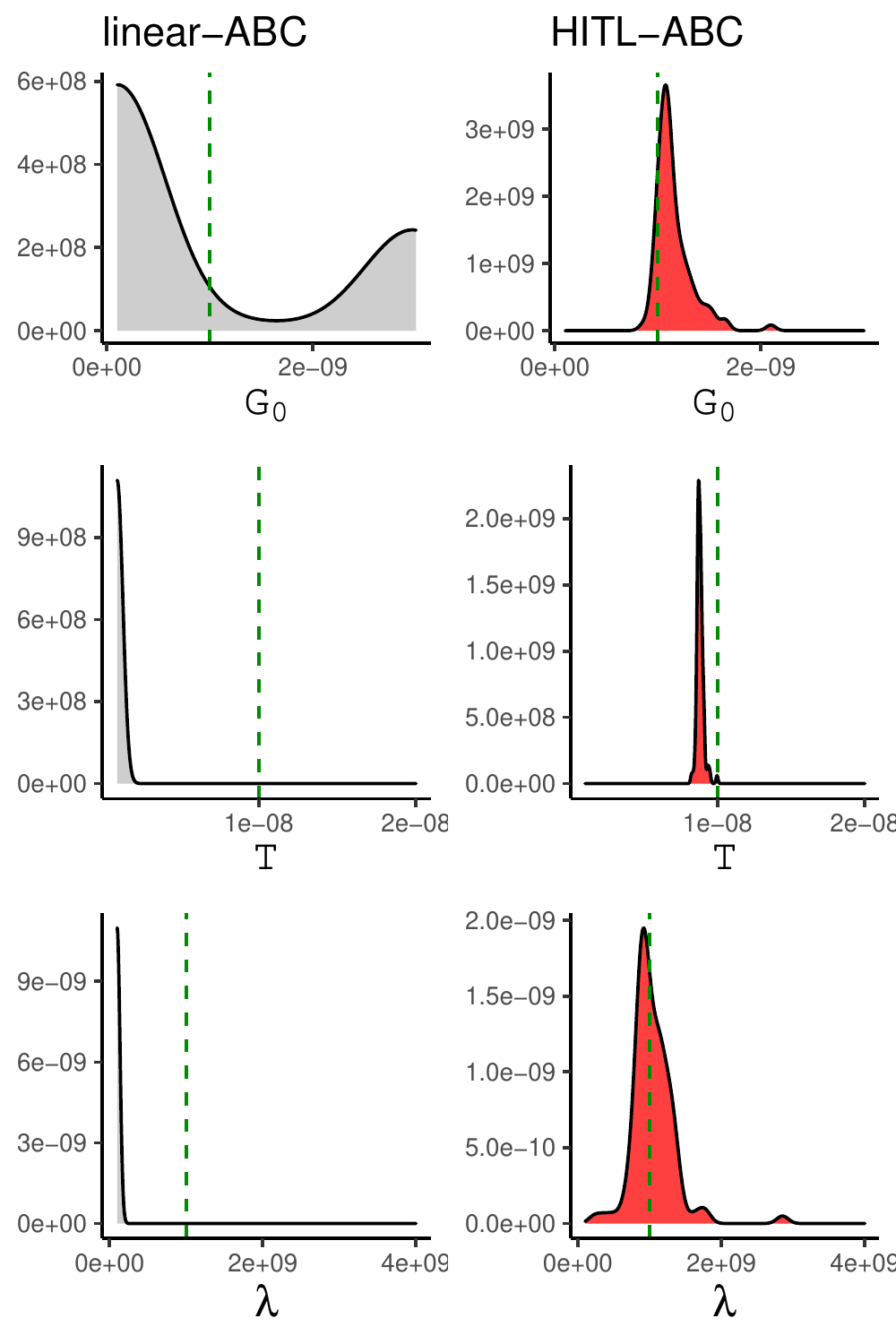} \\
            (a) $\zeta = 0$ & (b) $\zeta = 5$ & (c) $\zeta = 10$
        \end{tabular}
    \caption{HITL-ABC radically outperforms linear-ABC under model misspecification ($\zeta>0$, panels (b) and (c)). Approximate posteriors of the parameters of the radio propagation model obtained from HITL-ABC (red) and linear-ABC (grey) at varying levels of misspecification. The dashed green line denotes the true parameter value. For $\zeta=0$ (panel (a)), the model is correctly specified.  Prior is $\mathcal{U}([10^{-10}, 3\times 10^{-10}] \times [10^{-9}, 2\times 10^{-8}] \times [10^{8}, 4\times 10^{9}])$. Settings: $B=4\times 10^9$, $n_s = 801$, $\nobs = 300$, $\Nsim = 2000$, $\epsilon = 5\%$.}
    \label{fig:misspec}
    \end{center}
\end{figure*}
\subsection{Experiment under model misspecification}\label{sec:shotNoise}

We study the performance of the HITL-ABC method against that of linear-ABC \cite{Beaumont2002} under model misspecification. More precisely, we consider the challenging problem of estimating parameters of a stochastic radio channel model having intractable likelihood. Driven by an underlying point process, such models simulate radio propagation phenomena and are used to test and design wireless communication systems \cite{Goldsmith2005}. The potential of likelihood-free methods for inferring parameters of such models has been recognized recently \cite{Bharti2020, Ramoni2021}.

\paragraph{Data and model description} Radio channel data is measured in the frequency bandwidth $B$ at $n_s$ equidistant points, resulting in a frequency separation of $\Delta f = B/(n_s - 1)$. The measured transfer function data is $(Y_0, Y_1, \dots, Y_{n_s-1})$. The time-domain signal $y(t)$ is obtained by inverse Fourier transforming $\{Y_i\}_{i=0}^{n_s-1}$ to $y(t) = \frac{1}{n_s} \sum_{i=0}^{n_s-1} Y_i \exp(j2\pi i \Delta f t)$. Multiple realizations yield an $\nobs \times n_s$-dimensional data matrix. We focus on the model by \citet{Turin1972} who define the transfer function as $Y_i = \sum_l \alpha_l \exp(-j2\pi \Delta f i \tau_l)$, where $\tau_l$ is the time-delay and $\alpha_l$ is the complex gain of the $l^\textup{th}$ component. The delays are modeled as a one-dimensional Poisson point process with arrival rate $\lambda$. The gains $\alpha_l$, conditioned on $\tau_l$, are modeled as i.i.d. zero-mean circular complex Gaussian variables with conditional variance $\mathbb{E}[\vert \alpha_l \vert^2 | \tau_l] = G_0 \exp(-\tau_l/T)/ \lambda$. Therefore, the parameter vector constitutes $\theta = (G_0, T, \lambda)$. As the underlying points $\{\tau_l, \alpha_l\}$ are unobserved, the likelihood becomes intractable. The high dimensionality of the data compounds the issue, as $n_s$ can be of the order of a few thousands.

\begin{figure}[t]
    \begin{center}
    \includegraphics[width=\linewidth]{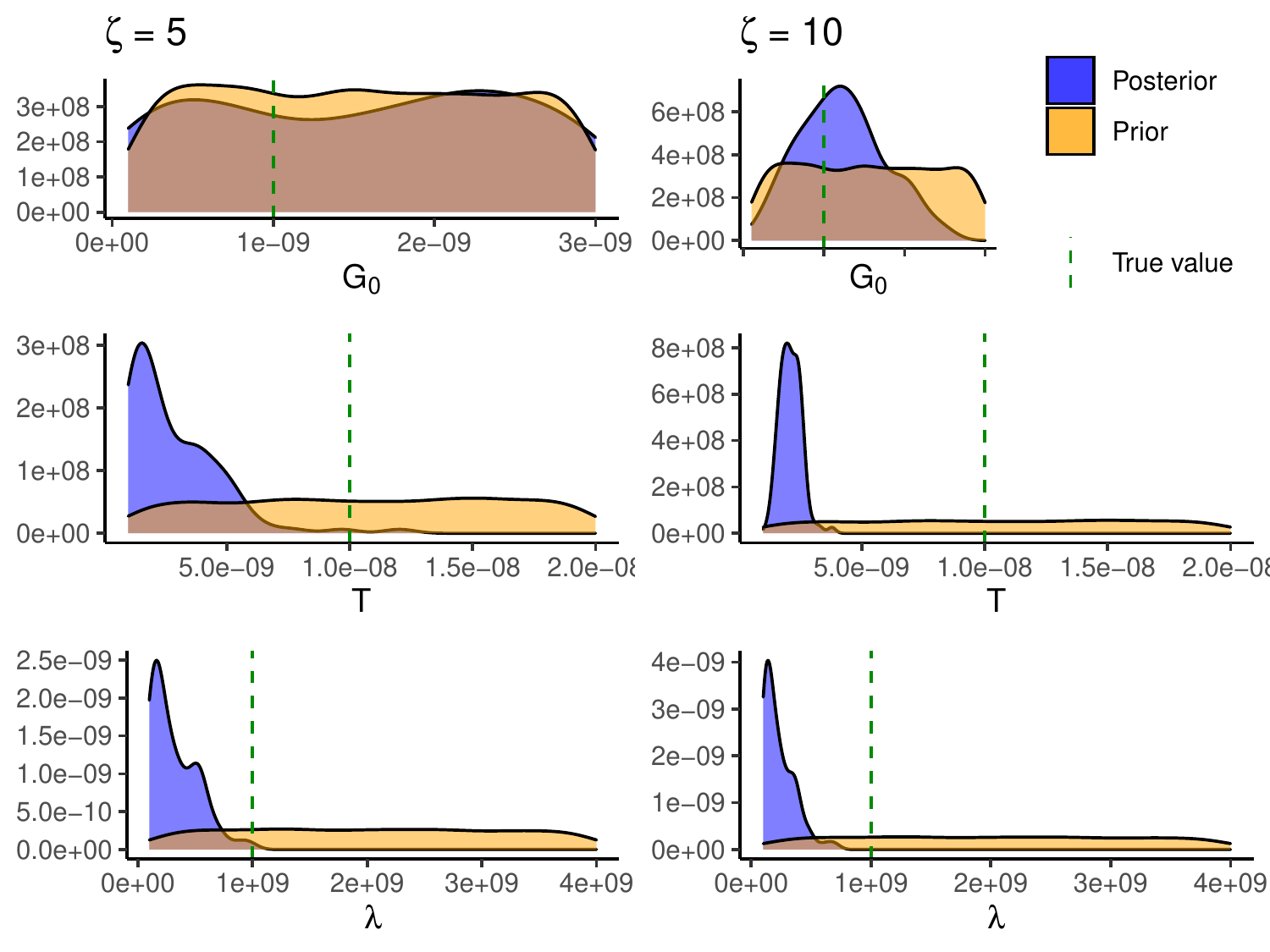}
    \caption{
    The expert is presented with the ABC posteriors before (yellow) and after (blue) including a statistic in the misspecification experiment. Figure shows the information provided to the expert for the statistic $\mathrm{var}(m_0)$. As $\mathrm{var}(m_0)$ is misspecified for $\zeta>0$,  the ABC posteriors for $T$ and $\lambda$ get concentrated far away from the true value, on the prior boundaries. Based on this information, the expert provides feedback on whether that statistic should be included.}
    \label{fig:expertView}
    \end{center}
\end{figure}

\paragraph{Setting} We consider the means and variances of the first three log-temporal moments $m$ as the summary statistics,
\begin{equation}\label{eq:temporalMoments}
    m_i = \log \int_0^{\frac{1}{\Delta f}} t^i \vert y(t)\vert^2 \text{dt} , \quad i = 0,1,2,
\end{equation}
as they have been shown to be informative about the parameters of interest \cite{Bharti2020, Ramoni2021}. Thus, the total number of statistics is $w=6$. To create a misspecified model, we perturb one of the temporal moments to produce a mismatch between observed and simulated statistics. Specifically, we compute observed statistics using $\theta_{\mathrm{true}} = (10^{-9}, 10^{-8}, 10^{9})$, and add zero-mean Gaussian random variables with variance $\zeta$ to $m_0$. This leads to $\mathrm{var}(m_0)$ being the only misspecified statistic. As a result, no setting of parameters yields temporal moments that match the observed values. We assess the performance of ABC methods by considering $\zeta = \{0,5,10\}$.

\paragraph{Expert involvement} In this experiment, we involved the expert to detect misspecification by showing them inference results. To that end, we asked for real feedback from a radio propagation expert. We first asked the expert to confirm the relevance of the statistics obtained from literature, prior to running the experiment. At each iteration, presenting just the utility maximizing statistic to the expert may not be sufficient for them to qualify it as being misspecified. Thus, the expert was also presented with the ABC posterior obtained before and after including the utility maximizing statistic. In particular, this gave the expert the opportunity to observe the impact of the statistic on the ABC posteriors, and potentially exclude it if they deem it to be misspecified.

\paragraph{Results} We observe in Fig.~\ref{fig:misspec}-a that when the model is correctly specified ($\zeta = 0$), both methods yield similar performance, as expected. When misspecification occurs ($\zeta > 0$), as shown in Fig.~\ref{fig:misspec}-b and Fig.~\ref{fig:misspec}-c, the performance of linear-ABC seriously degrades --- posterior samples become concentrated further away from the correct value, on the prior boundary for $T$ and $\lambda$. The posterior of $G_0$ is also hampered significantly. That is not the case for HITL-ABC, as the expert involved is able to observe the effect of the $\mathrm{var}(m_0)$ statistic on the ABC posterior (as shown in Fig.~\ref{fig:expertView}) and exclude it from being selected. Hence, the performance of HITL-ABC remains relatively stable as the level of misspecification increases. Additional results of the experiment can be found in Appendix~\ref{app:misspec}. 

\subsection{Experiment in low-simulation regime}\label{sec:gk}

We now compare the performance of the proposed HITL-ABC method against linear-ABC \cite{Beaumont2002}, neural-ABC \cite{Blum2010nonlinear}, and ridge-ABC \cite{Blum2013} in low-simulation regimes. We also include the statistics selection method of \citet{Barnes2012} (implemented with $\delta=0.1$) combined with linear-ABC for comparison. We demonstrate the results using the g-and-k distribution \cite{Prangle2020}, which is a flexible univariate distribution without a closed-form density. It is defined by its inverse cumulative distribution function
\begin{multline}
    F^{-1}(x;A,B,c,g,k) =\\ A + B\left[ 1 + c \frac{1 - \exp{(-g z(x))}}{1 + \exp{(-g z(x))}} \right] (1 + z(x)^2)^k z(x), \nonumber
\end{multline}
where $z(x)$ is the $x^\textup{th}$ standard Gaussian quantile. Keeping $c=0.8$ fixed \cite{Rayner2002}, the unknown parameters $\theta = (A,B,g,k)$ govern the location, scale, skewness, and kurtosis of the distribution, respectively. 

\paragraph{Setting} The pool of statistics consists of estimates of these quantities: $s_A = L_2$, $s_B = L_3 - L_1$, $s_g = L_3 + L_1 - 2L_2)/s_B$, and $s_k = (E_7 - E_5 + E_3 - E_1)/s_B$ where $L_i$ and $E_j$ are the $i^\textup{th}$ quartile and $j^\textup{th}$ octile, respectively \cite{Drovandi2011}. We also include pairwise products of these four statistics and five uniform random variables $u_i \sim \mathcal{U}(0,1)$, $i=1,\dots,5$ in $\mathcal{S}$, yielding a total of $w=15$ statistics. The expert feedback is simulated using Eq.~$\eqref{eq:feedbackModel}$ with $\mathbf{s}^* = \{s_A, s_B, s_g, s_k\}$. The priors for all the parameters are set to $\mathcal{U}(0,10)$. The true parameter value is $\theta_{\mathrm{true}} = (3,4,2,1)$. The statistics are computed using $\nobs = 10,000$ data points from the g-and-k distribution. We vary the simulation budget $\Nsim$, and run the ABC methods 100 times for each $\Nsim$ with $\epsilon=10\%$ (meaning the available simulations are different for each run). The accuracy of the obtained ABC posteriors is assessed by estimating the KL divergence between them and a reference ABC posterior using Eq.~\eqref{eq:KL_estimator}, as the likelihood is intractable. The reference ABC posterior is obtained using linear-ABC with $\Nsim = 10,000$ and $\epsilon = 1\%$. 

\begin{figure*}[t]
\begin{center}
\includegraphics[width=0.33\linewidth]{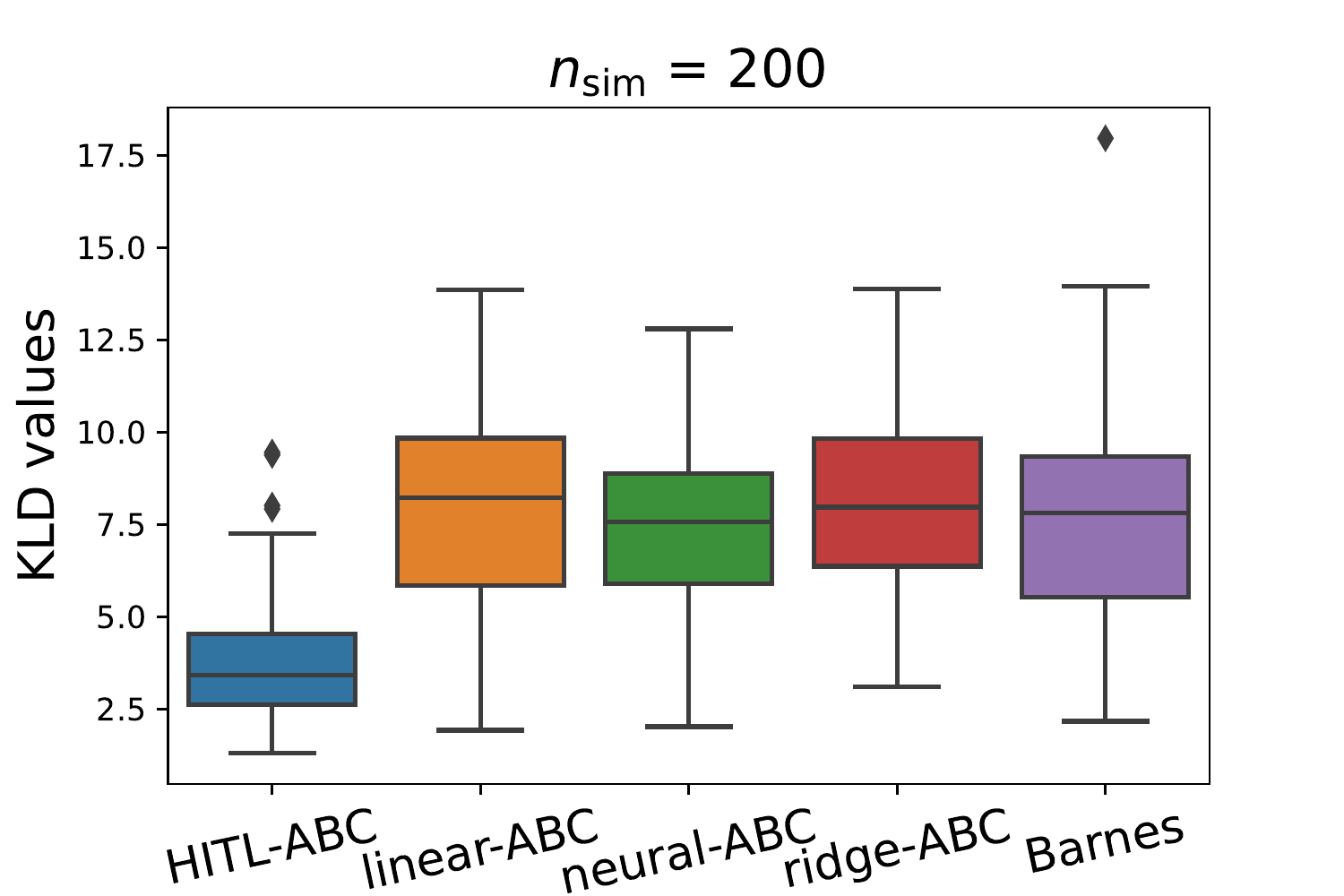}
\includegraphics[width=0.33\linewidth]{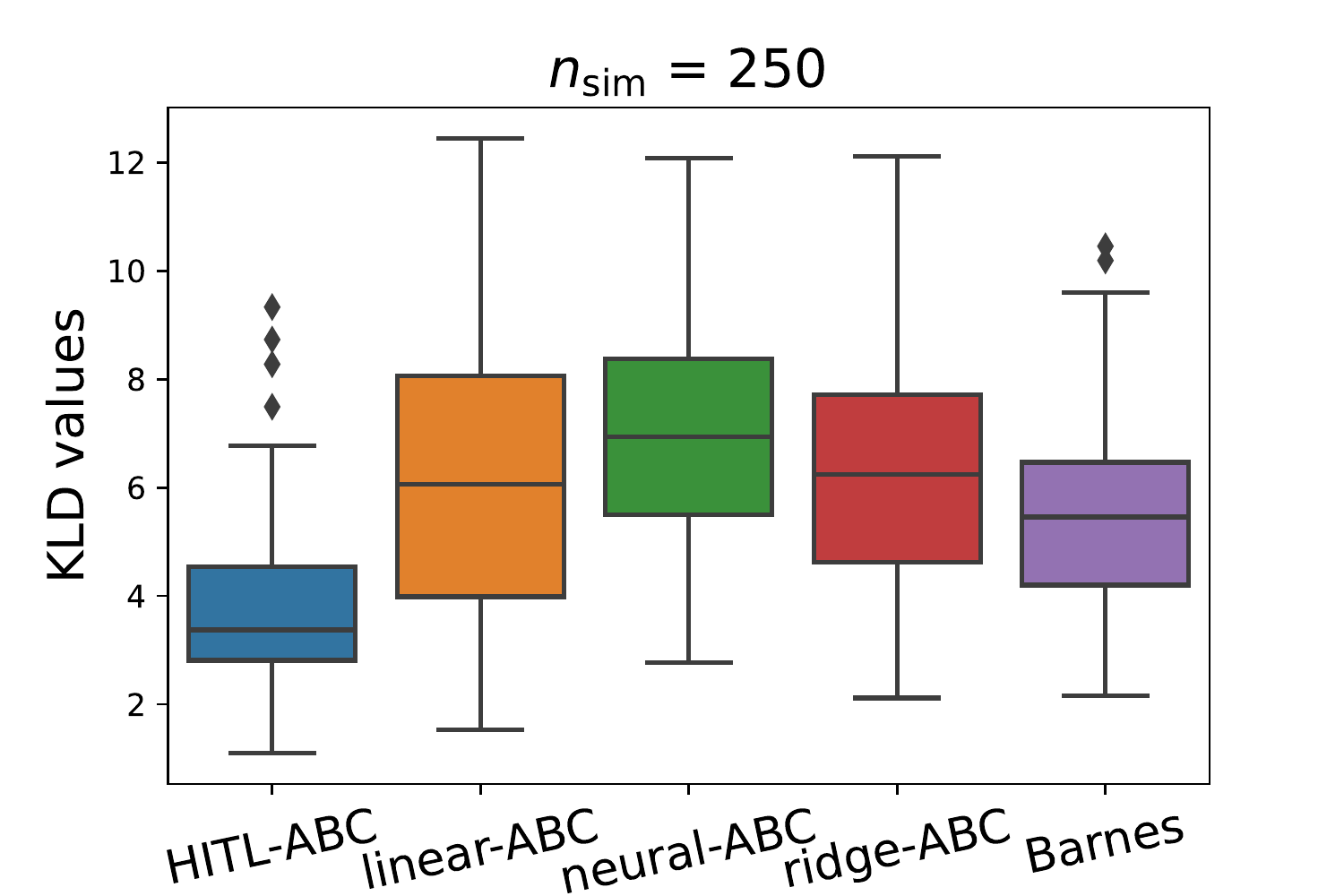}
\includegraphics[width=0.33\linewidth]{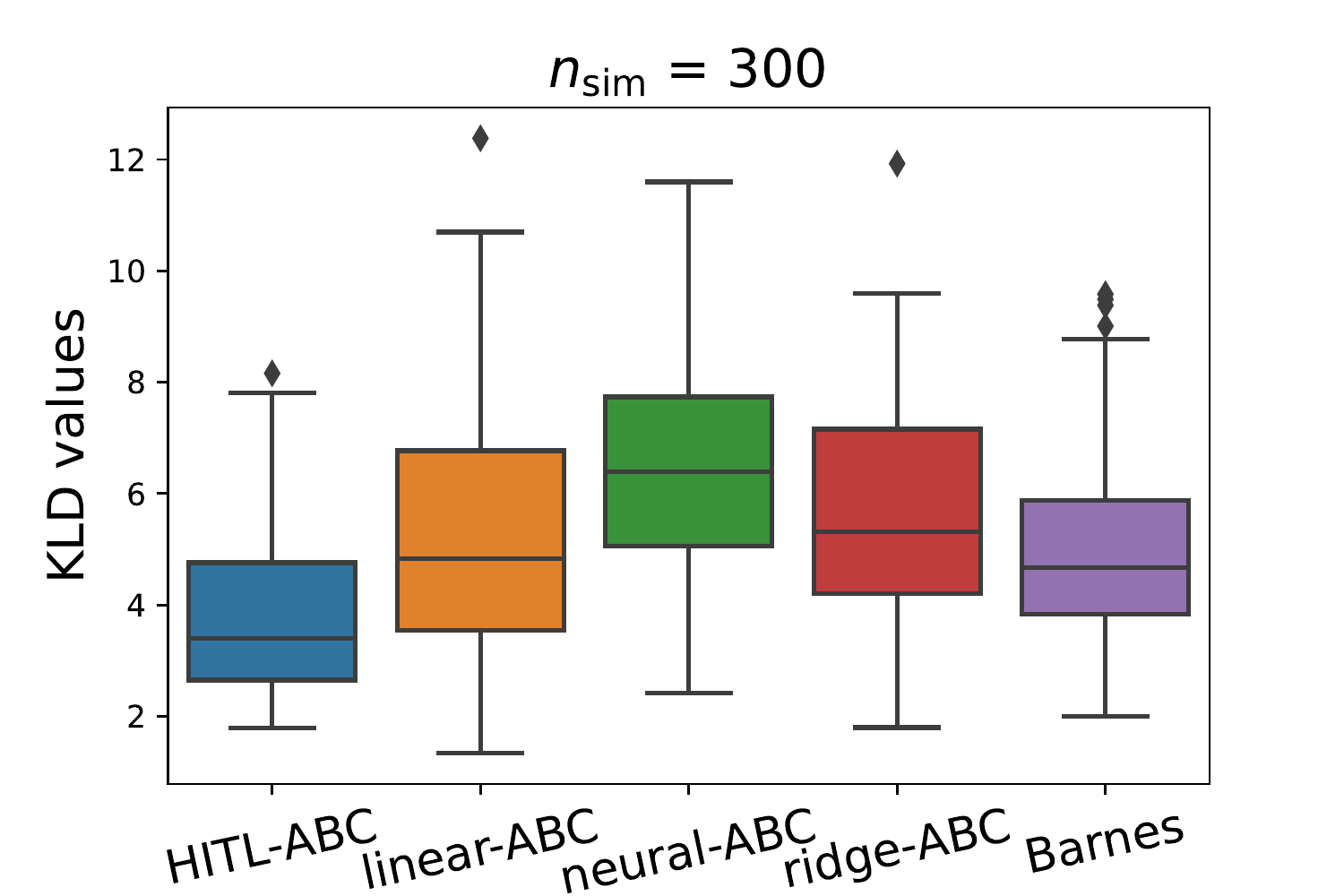}
\includegraphics[width=0.33\linewidth]{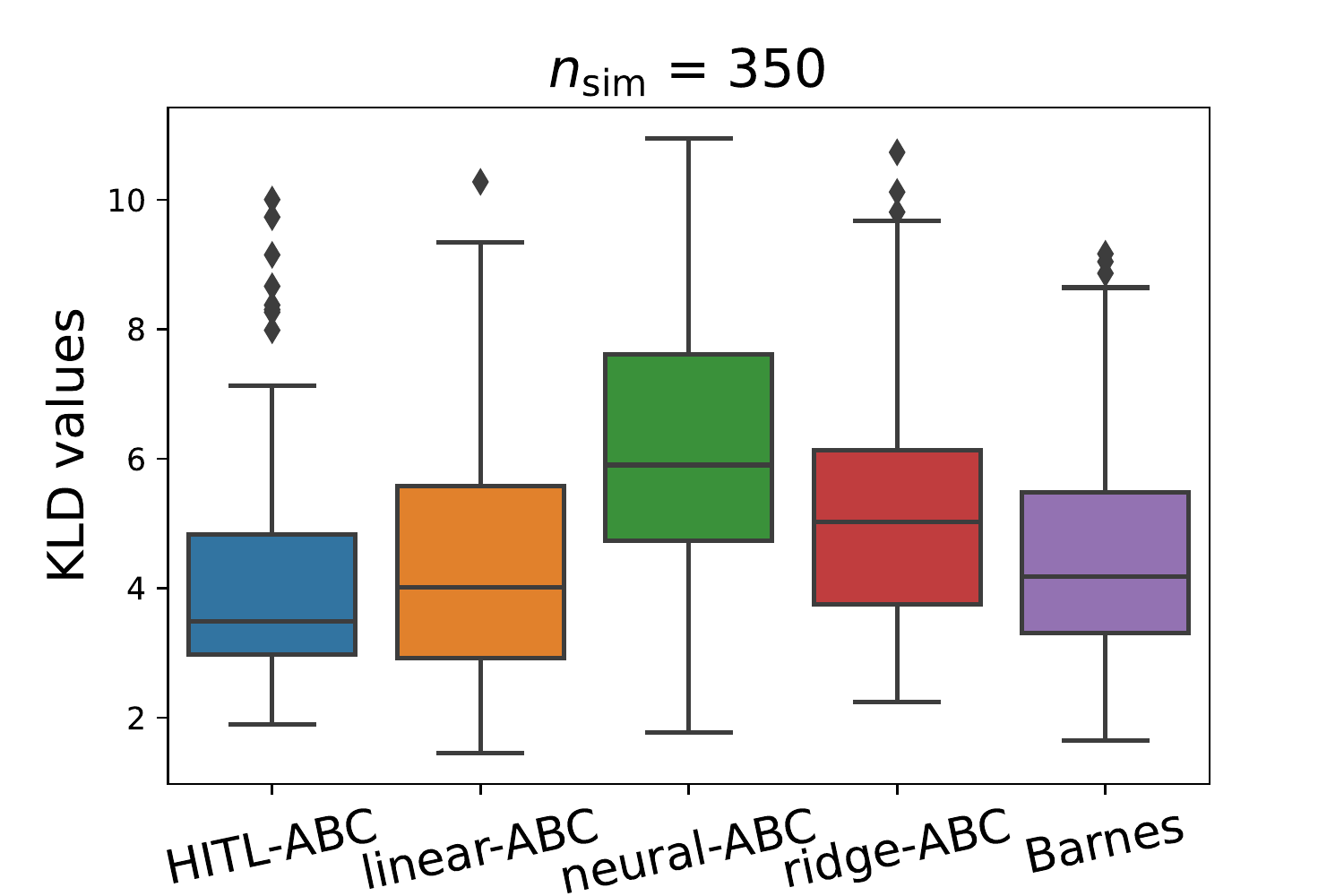}
\includegraphics[width=0.33\linewidth]{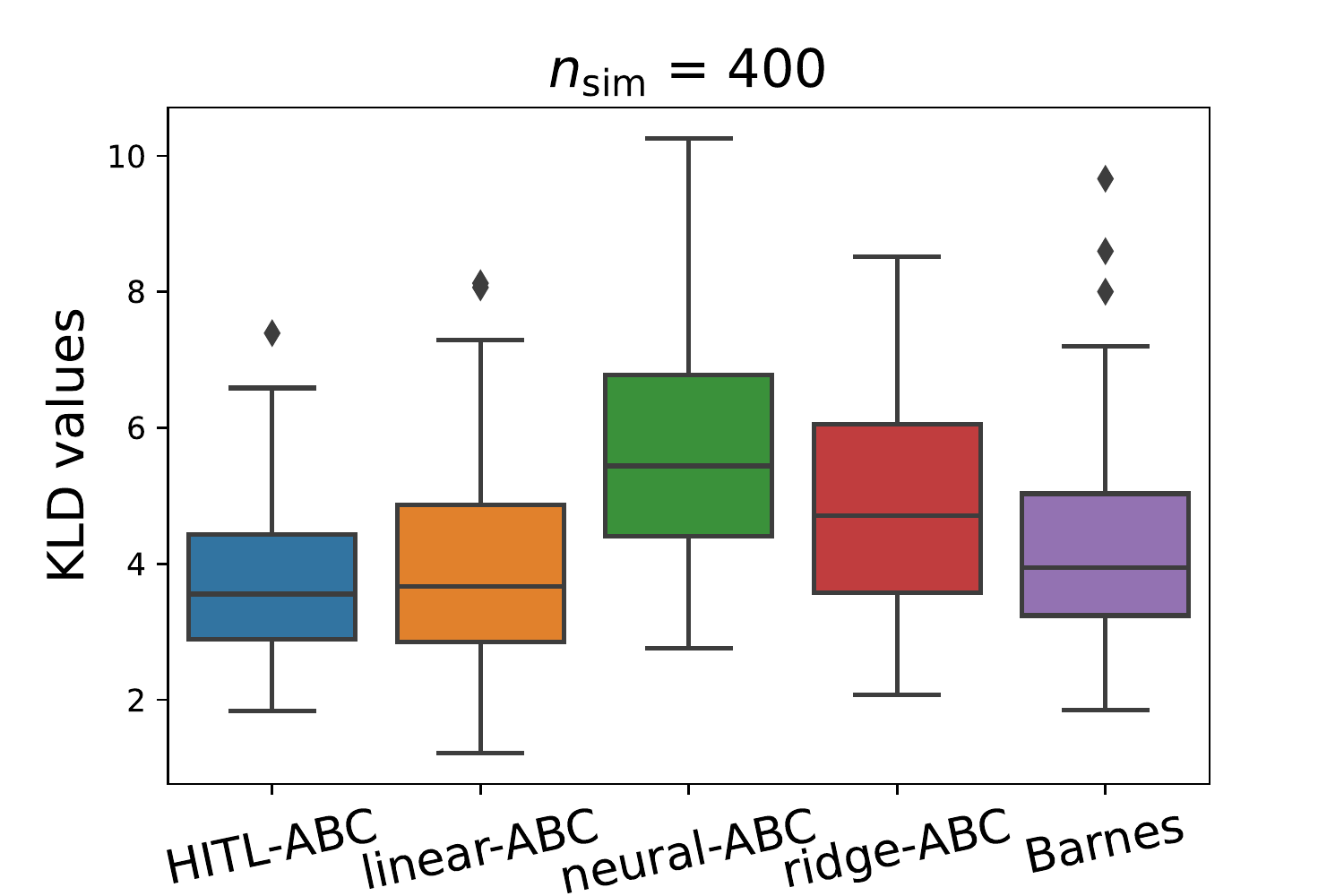}
\includegraphics[width=0.33\linewidth]{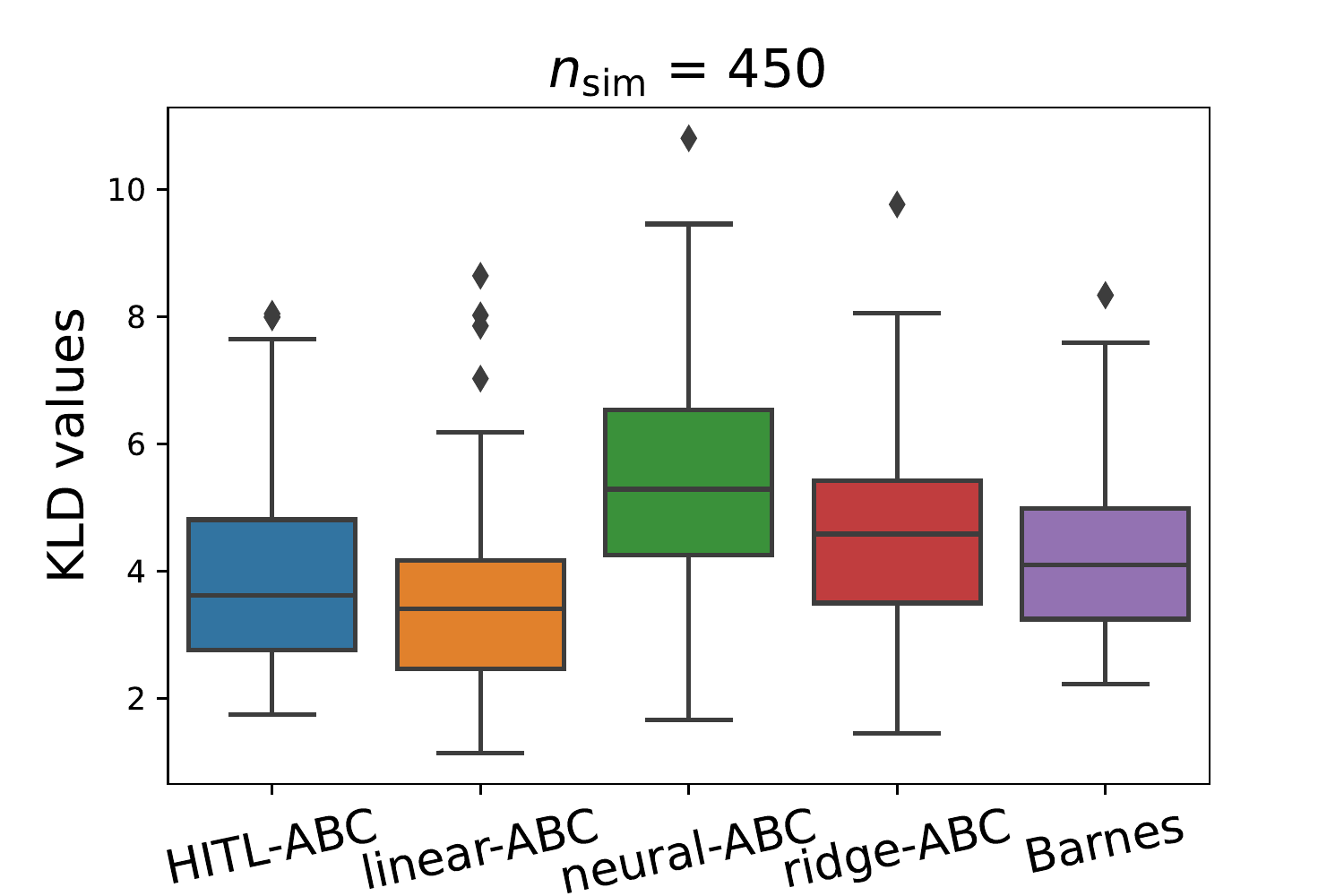}
\caption{The proposed HITL-ABC outperforms the other regression-ABC methods which do not involve experts, on low-simulation regimes ($\Nsim \leq 350$) and  performs on-par with larger numbers of simulations. Box plots of KL divergence values between ABC posteriors from different methods at varying $\Nsim$ and a reference ABC posterior obtained with $\Nsim=10,000$. Lower values of KL divergence indicate better posterior characterization.}
\label{fig:gk}
\end{center}
\end{figure*}

\paragraph{Results} The results are shown in Fig.~\ref{fig:gk}. We observe that the proposed HITL-ABC method outperforms the various regression-ABC methods along with Barnes' method for low values of simulation budget $\Nsim$. For $\Nsim = 400$ and above, the performance of HITL-ABC and linear-ABC is at par. As $\Nsim$ increases, the KL divergence values of regression-ABC methods decrease, indicating improved performance. Amongst the regression-ABC methods, linear-ABC performs the best, followed by ridge-ABC and neural-ABC for most values of $\Nsim$.

For low values of $\Nsim$, KL divergence estimates exhibit larger variance, leading to wrongly selecting non-informative statistics in the Barnes' method. However, in HITL-ABC, the lack of available simulations is compensated by the expert feedback, resulting in similar KL divergence values for each $\Nsim$. This can be seen from Table~\ref{tab:expertLoad}, where the average number of required feedback increases as $\Nsim$ decreases. Moreover, we see that maximizing the utility in Eq.~\eqref{eq:utility} is more efficient in terms of yielding the least number of feedback than a random query acquisition strategy.

\begin{table}[]
\centering
\caption{Average number of expert feedback required in the low-simulation regime experiment as a function of simulation budget.}
    \begin{tabular}{ccccccc}
    \toprule
    $n_{\mathrm{sim}}$ & 200  & 250 & 300 & 350 & 400 & 450 \\ \midrule
    HITL-ABC & \textbf{10.1} & \textbf{8.5} & \textbf{8.3} & \textbf{6.3} & \textbf{6.0} & \textbf{6.3} \\
    Random & 13.8 & 13.6 & 13.4 & 13.3 & 13.1 & 13.4 \\ \bottomrule
    \end{tabular}
\label{tab:expertLoad}
\end{table}

\begin{table}[]
\centering
\caption{Number of times the optimal set of summary statistics, i.e., just the sample mean and sample variance, were selected out of 100 runs by the HITL-ABC, for varying values of hyperparameters $\pi$ and $\rho$.}
\begin{tabular}{@{}rl | rl@{}}
\multicolumn{2}{c}{$\rho = 0.5$}                                            & \multicolumn{2}{c}{$\pi = 0.95$}                                             \\ \midrule
\multicolumn{1}{c}{$\pi$} & \multicolumn{1}{c}{$(\hat \mu, \hat \sigma^2)$} & \multicolumn{1}{c}{$\rho$} & \multicolumn{1}{c}{$(\hat \mu, \hat \sigma^2)$} \\ \midrule
1.0                       & 100\%                                               & 0.2                        & 92\%                                                \\
0.95                      & 89\%                                                & 0.3                        & 91\%                                                \\
0.9                       & 72\%                                                & 0.4                        & 91\%                                                \\
0.85                      & 70\%                                                & 0.6                        & 94\%                                                \\
0.8                       & 50\%                                                & 0.7                        & 90\%                                                \\
0.75                      & 27\%                                                & 0.8                        & 95\%                                                \\ \bottomrule
\end{tabular}
\label{tab:hyperparameter}
\end{table}

\subsection{Sensitivity to hyperparameter setting}\label{sec:hyperparameter}
\paragraph{Setting} Finally, we perform a hyperparameter sensitivity analysis on a toy problem of estimating the parameters $\theta = (\mu, \sigma^2)$ of a Gaussian distributed random variable $y_1,\dots,  y_{n_{\mathrm{obs}}} \sim \mathcal{N}(\mu, \sigma^2)$. In this case, the sample mean $\hat \mu$ and the sample variance $\hat \sigma^2$ are sufficient statistics for inferring $\theta$. Additionally, we include the range ($\text{max}_i ~y_i - \text{min}_i ~y_i$) and two uninformative statistics $u_1, u_2 \sim \mathcal{U}(0,1)$ in the pool of statistics, i.e., $\mathcal{S} = \{\hat \mu, \hat \sigma^2, \text{range}, u_1, u_2\}$. We set the true parameter value to $\theta_{\mathrm{true}} = (0,2)$ and prior to $\mathcal{U}([-5,5]\times [0,5])$. The ABC method is run with $\nobs =  500$, $\Nsim = 2000$, and $\epsilon=5\%$.

\paragraph{Results} We vary the values of $\rho$ and $\pi$, and report the number of times only the sufficient statistics $(\hat \mu, \hat \sigma^2)$ are selected out of 100 runs in Table~\ref{tab:hyperparameter}. Firstly, we observe that HITL-ABC is able to pick the sufficient statistics each time in case of a noiseless  feedback ($\pi=1$). As expected, when the value of $\pi$ decreases, the sufficient statistics are picked less often. Additionally, we observe that varying $\rho$ barely has any effect on the output of the algorithm. Finally, keeping $\pi = 0.95$ and $\rho = 0.5$ fixed, we vary the value of the stopping criterion $\delta$. We report the average number of queries to the expert over $100$ runs in Table~\ref{tab:delta}. As the value of $\delta$ increases, the average number of feedback decreases. This could also serve as a rule of thumb on how to set $\delta$, which could depend on how much the expert wishes to be involved, i.e., the maximum number of times they want to be queried.

\begin{table}[]
    \centering
    \caption{Average number of expert feedback required in the Gaussian example w.r.t. the stopping criterion $\delta$.}
    \begin{tabular}{cc}
    \toprule
    $\delta$ & Avg. no. of feedback. \\
    \midrule
    0.02 & 3.04 \\
    0.04 & 2.49 \\
    0.06 & 2.24 \\
    0.08 & 2.18 \\
    0.10 & 2.17 \\
    \bottomrule
    \end{tabular}
    \label{tab:delta}
\end{table}

\section{Conclusion}

In this paper, we introduced the first ABC method that actively leverages domain knowledge from experts in order to select summary statistics. Involving the experts in the ABC method gives us the opportunity to handle misspecified models, something the existing methods fail in.
With fairly limited effort from the expert (answering yes/no when presented with a few statistics), we are able to outperform the regression-ABC methods in situations where the simulation budget is low. This simple binary feedback could potentially be scaled to include multiple experts, albeit at the cost of added complexity to determine which expert to ask feedback from. The method also acts as an assistant for the experts to try out different statistics without much effort, however, the usefulness of this method as an AI assistant is a topic for future studies. Finally, there is avenue for further research on extending other likelihood-free inference methods to be amenable to expert's feedback.

\paragraph{Limitations} Our method inherits the limitations of all the greedy statistics selection ABC methods, i.e., 1) due to the step-wise selection approach adopted in our method, there is no guarantee of converging to the best subset of statistics, and the method may only converge to a local optimum; and 2) applying it in combination with computationally expensive ABC methods such as ABC-MCMC \cite{Marjoram2003} or ABC-SMC \cite{Beaumont2009} can be infeasible. Lastly, it might be tempting to propose eliciting feedback about a statistic by always showing the posteriors before and after including it. However, that runs the risk that the users may amplify noise in the statistics, especially in low-simulation regimes, if they are not careful. Using so-called ``posterior elicitation'' and inferring the priors indirectly may then be helpful \cite{Daee2018}.

\section*{Acknowledgements}

This work was supported by the Academy of Finland (Flagship programme: Finnish Center for Artificial Intelligence FCAI). SK was also supported by the UKRI Turing AI World-Leading Researcher Fellowship, EP/W002973/1. We acknowledge the computational resources provided by the Aalto Science-IT Project from Computer Science IT.

\bibliography{bibliography}
\bibliographystyle{icml2022}

\newpage
\appendix
\onecolumn

{
\begin{center}
\Large
    \textbf{Supplementary Materials}
\end{center}
}

\section{Background on Bayesian Experimental Design} \label{app:bed}

Experimental design tackles the question of selecting the most informative experimental design $d \in \mathcal{D}$ to learn about a parameter $\theta$. To this end, we must choose a so-called utility function $U : \mathcal{D} \rightarrow \mathbb{R}$ which assesses the worth of design $d$, and the optimal design is then
\begin{equation}
     d^{\star} = \argmax_{d \in \mathcal{D}} U(d).
\end{equation}

Let us assume that experimental design $d$ leads to observation $y$. In \emph{Bayesian} experimental design \citep{chaloner1995bayesian,ryan2016review}, we are equipped with a probabilistic model $p(y|\theta,d)$ as well as a prior distribution for the parameter of interest $p(\theta)$. A principled utility function from an information-theoretic perspective is the expected Kullback-Leibler (KL) divergence between the future posterior $p(\theta|d,y)$ and the current prior distribution $p(\theta)$:
\begin{equation}
     U(d) = \mathbb{E}_{p(y|d)} \left[ \text{KL}(p(\theta|d,y) || p(\theta) ) \right].
    \label{eq-eig}
 \end{equation}
This utility function can equivalently be presented as the so-called expected information gain, that is, the expected reduction in (differential) entropy from the prior to the posterior distributions. Another equivalent definition is the mutual information between $y$ and $\theta$ given design $d$ (usually denoted $\text{I}(y;\theta|d)$). A closed-form expression of Eq.~\eqref{eq-eig} is not available in general, and a common estimation strategy consists of Monte Carlo (MC) approximation, which is more precisely a nested MC approximation \citep{rainforth2018nesting}.

Lastly, we mention that BED is usually applied in a (myopic) sequential way. This means that once the optimal design has been found, and the associated experiment has been run, we proceed to update the posterior distribution of $\theta$, which now acts as the prior distribution for the next step. The new utility is optimized again, and so on and so forth. Experiments are thus run one-by-one. Formally, at iteration $k+1$, if previously obtained designs $d_1, \dotsc d_k$ led to observations $y_1, \dotsc, y_k$, respectively, we have
\begin{align}
  U_{k+1}(d) = & \mathbb{E}_{p(y|d,d_{1:k}, y_{1:k})}  \left[ \text{KL}(p(\theta|d,y,d_{1:k}, y_{1:k}) || p(\theta|d_{1:k}, y_{1:k}) ) \right]. \notag 
 \label{eq-eig-seq}
\end{align}

\section{Additional Results of Misspecification Experiment} \label{app:misspec}

\begin{figure*}[h]
    \begin{center}
        \begin{tabular}{ccc}
            \includegraphics[width=0.49\linewidth]{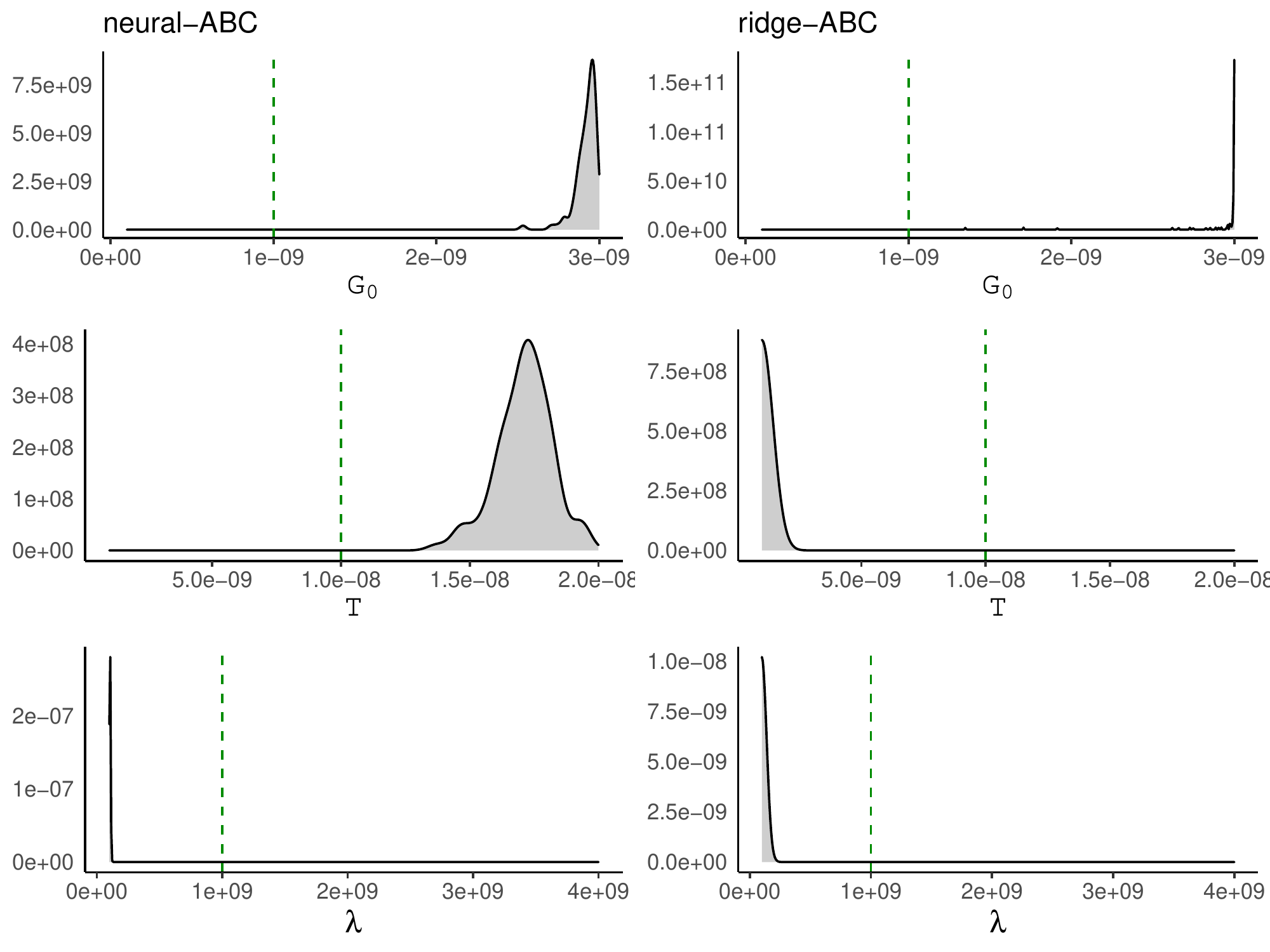} & 
            \includegraphics[width=0.49\linewidth]{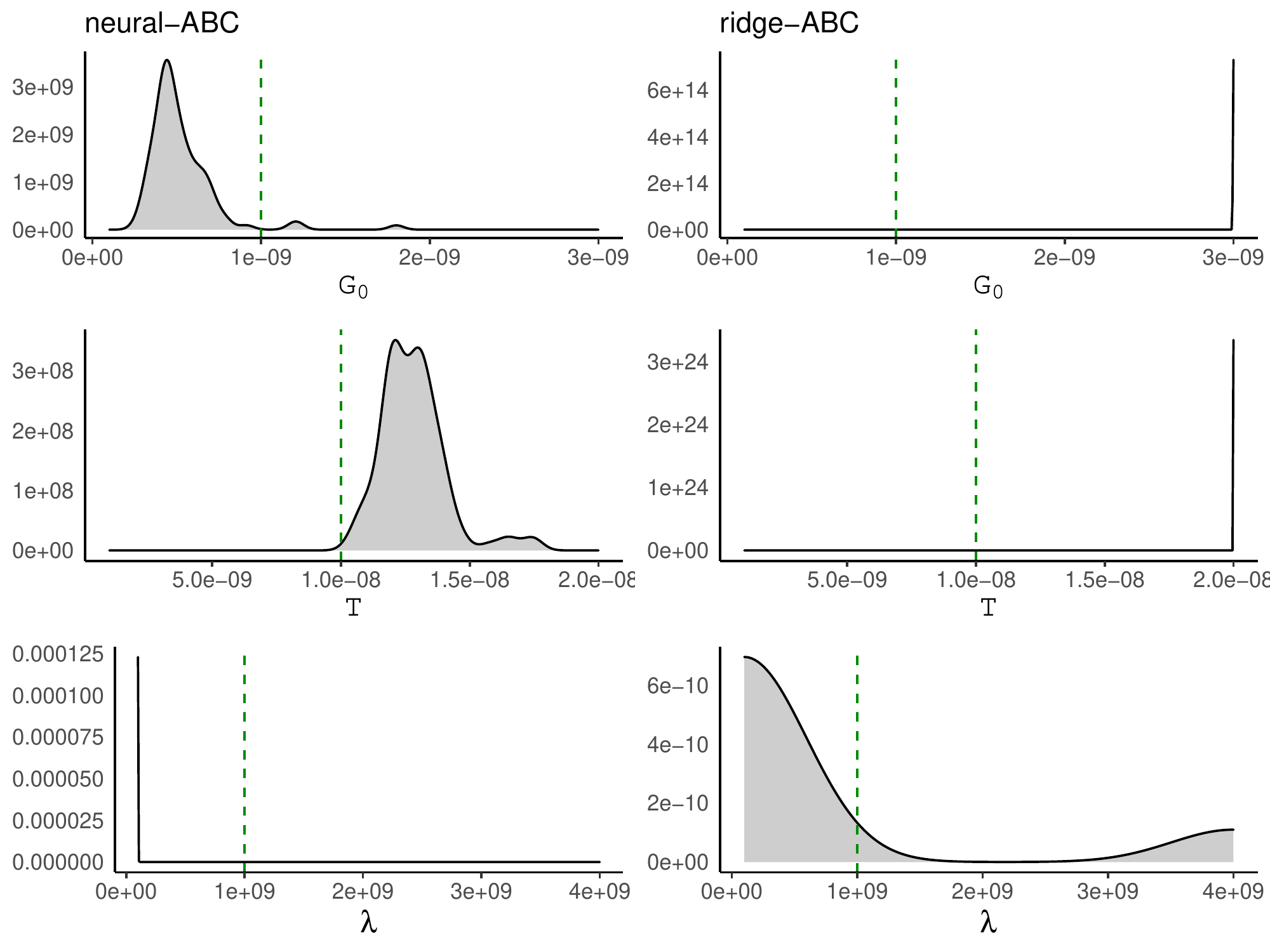}  \\
              (a) $\zeta = 5$ & (b) $\zeta = 10$
        \end{tabular}
    \caption{Approximate posteriors of the parameters of the radio propagation model obtained from neural-ABC and ridge-ABC at varying levels of misspecification. The dashed green line denotes the true parameter value. Prior is $\mathcal{U}([10^{-10}, 3\times 10^{-10}] \times [10^{-9}, 2\times 10^{-8}] \times [10^{8}, 4\times 10^{9}])$. Settings: $B=4\times 10^9$, $n_s = 801$, $\nobs = 300$, $\Nsim = 2000$, $\epsilon = 5\%$.}
    \label{fig:misspec_neural_ridge}
    \end{center}
\end{figure*}

In this section, we present additional results on the misspecification experiment conducted on the Turin model \cite{Turin1972}. In Fig.~\ref{fig:misspec_neural_ridge}, we show the approximate posteriors obtained from neural-ABC and ridge-ABC under model misspecification. Their results are similar to the one obtained from linear-ABC in Fig.~\ref{fig:misspec}, as expected, indicating the failure of all regression ABC methods in handling misspecified scenarios.

\begin{figure*}[t]
    \begin{center}
        \begin{tabular}{ccc}
            \includegraphics[width=0.7\linewidth]{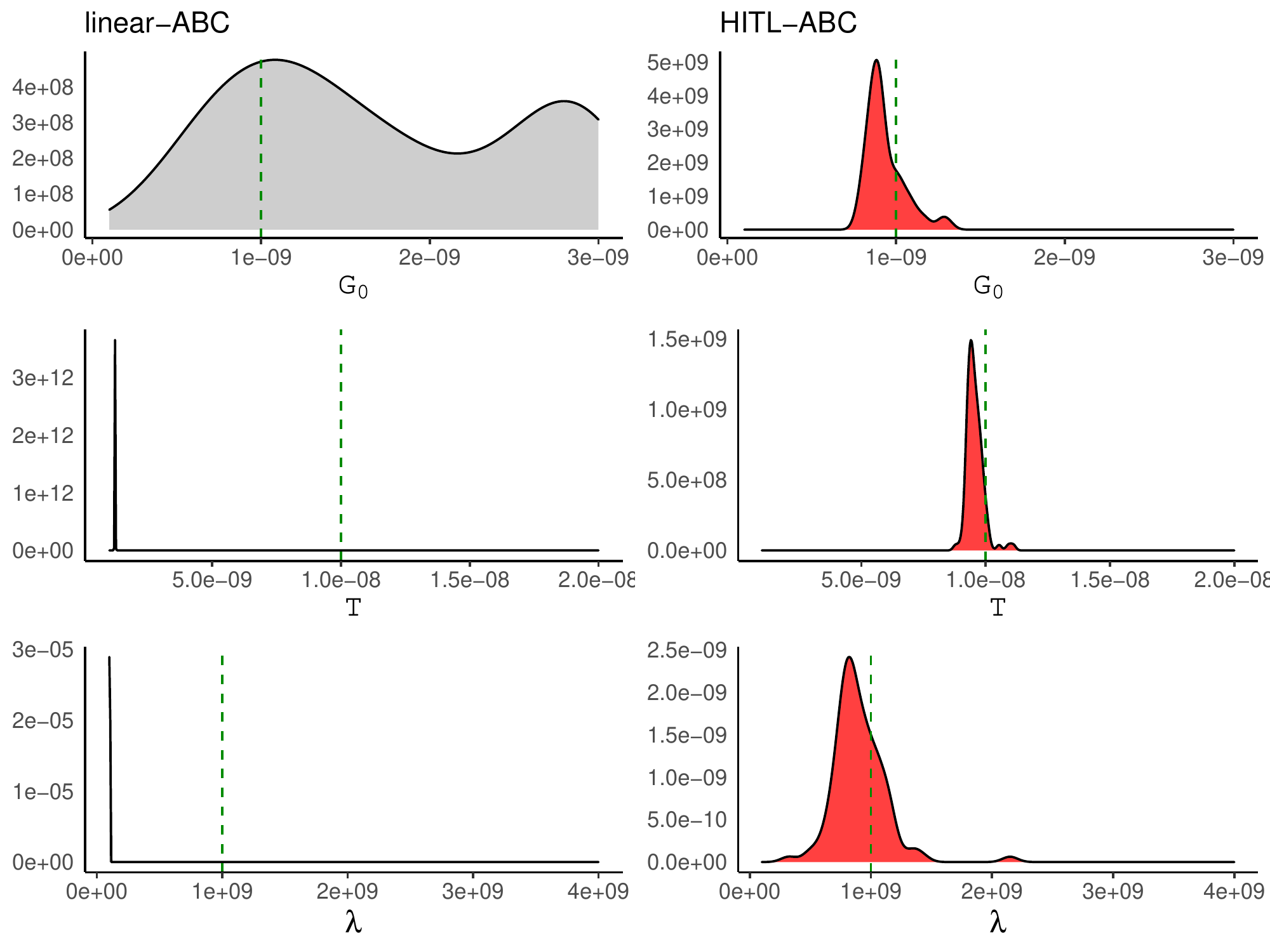}
        \end{tabular}
    \caption{Approximate posteriors of the parameters of the radio propagation model obtained from linear-ABC and HITL-ABC at $\zeta=1$. The dashed green line denotes the true parameter value. Prior is $\mathcal{U}([10^{-10}, 3\times 10^{-10}] \times [10^{-9}, 2\times 10^{-8}] \times [10^{8}, 4\times 10^{9}])$. Settings: $B=4\times 10^9$, $n_s = 801$, $\nobs = 300$, $\Nsim = 2000$, $\epsilon = 5\%$.}
    \label{fig:misspec_zeta1}
    \end{center}
\end{figure*}

We also include the results for $\zeta = 1$ in Fig.~\ref{fig:misspec_zeta1}, to demonstrate that even a small degree of misspecification leads to failure in linear-ABC method. On the other hand, HITL-ABC achieves better performance as the misspecified statistic is excluded by the expert. However, we remark that with lower levels of misspecification, it may become difficult for the expert to determine that a statistic is misspecified on seeing the inference results at each iteration of the sequential experiment. 

\end{document}